%% file: main.tex
\documentclass[10pt]{article} 
\usepackage[accepted]{tmlr}
\input{math_commands.tex}

\usepackage[dvipsnames]{xcolor} 
\usepackage{multirow}
\usepackage{hyperref}
\hypersetup{
    colorlinks=true,
    citecolor=Blue, 
    linkcolor=Blue, 
    urlcolor=Blue
}
\usepackage{url}
\usepackage[table]{xcolor}
\usepackage{colortbl}
\usepackage{microtype}
\usepackage{graphicx}
\usepackage{subcaption}
\usepackage{booktabs}
\usepackage{hyperref}

\usepackage{amsmath}
\usepackage{amssymb}
\usepackage{mathtools}
\usepackage{amsthm}
\usepackage[capitalize,noabbrev]{cleveref}
\usepackage{multirow} 
\usepackage{rotating} 

\definecolor{darkgreen}{rgb}{0.0, 0.4, 0.0}

\definecolor{firebrick}{rgb}{0.698, 0.133, 0.133}

\title{Optimization as a Dynamical System: Generative Schedules from Latent ODEs}
\author{\name Matt L. Wiemann$^{*}$ \email matt.sampson@princeton.edu \\
      \addr Princeton University
      \AND
      \name Peter Melchior$^{*}$ \email peter.melchior@princeton.edu \\
      \addr Princeton University}
\footnotetext[1]{$^*$Equal contribution.}

\def\month{08}  
\def\year{2026} 
\def\openreview{\url{https://openreview.net/forum?id=SwmzJgB9TA}} 

\usepackage{arydshln}

\definecolor{lightgray}{gray}{0.7}
\arrayrulecolor{black}
\theoremstyle{plain}

\theoremstyle{definition}

\theoremstyle{remark}

\usepackage[textsize=tiny]{todonotes}

\usepackage{amsmath}
\usepackage{amssymb}
\usepackage{mathtools}
\usepackage{amsthm}

\usepackage{enumitem}
\usepackage{url}
\usepackage{graphicx}
\usepackage{algorithm}
\usepackage{algpseudocode}
\usepackage{pifont} 
\usepackage{tikz}
\usepackage{wrapfig}
\usepackage{comment}

\usepackage{booktabs,capt-of} 
\usepackage[capitalize,noabbrev]{cleveref}

\begin{document}
\maketitle

\begin{abstract}
We present a new meta-learning method to determine the optimal learning rate schedule for gradient descent. It leverages training runs from a hyperparameter search to learn a latent representation of the training process, which is modeled as a dynamical system. Given current training metrics, it predicts the future learning rate schedule with the best long-term validation performance.
Our scheduler generalizes beyond previously observed training dynamics and creates specialized schedules that deviate noticeably from even the best-performing parametric functions. 
It outperforms all baselines we compare to on results for image classification with CNN and ResNet models as well as for next-token prediction with a transformer model.
The trained models are located in flatter regions of the loss landscape and thus provide better generalization than those trained with other schedules.
Our method is computationally efficient, optimizer-agnostic, and can easily be layered on top of ML experiment-tracking platforms to streamline training of neural networks from scratch.
\end{abstract}

\begin{figure}[h]
    \centering
    \includegraphics[width=0.7\linewidth]{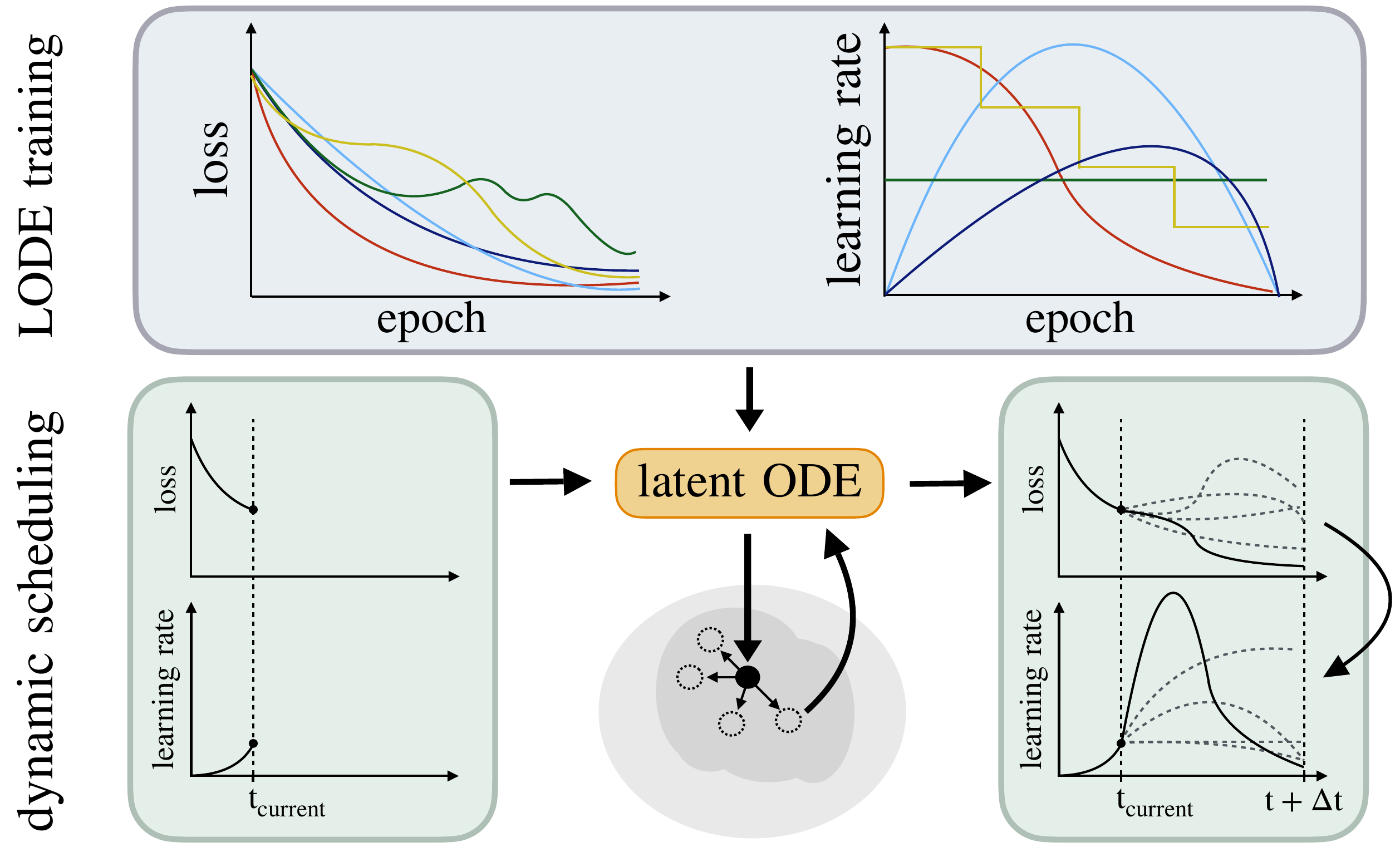}
    \caption{\small Schematic of the dynamic learning rate scheduler. \textbf{Latent ODE model:} Metrics tracked during a hyperparameter sweep contain a variety of learning rate schedules as well as resulting training and validation losses. We train a latent ODE model to reconstruct these training time series. \textbf{Dynamic scheduling:} For a new training run, we encode the current loss and learning rate time series with the latent ODE model, sample an ensemble in latent space centered on the current position, generate their future training behavior, and select the learning rate schedule with the best final validation performance.}
    \label{fig:lode-schematic}
\end{figure}

\section{Introduction}
Deep learning owes much of its success to the surprising effectiveness of gradient descent for high-dimensional, non-convex optimization. The simple update rule
\begin{equation}
    \boldsymbol{\theta}_{t+1} = \boldsymbol{\theta}_t - \eta_t \mathbf{g}_t
\end{equation}
for the network parameters $\boldsymbol{\theta}$ and the gradient $\mathbf{g}$ (possibly after averaging, clipping, or scaling) is easy to compute, but immediately begs the question: how to set the learning rate $\eta$ and its possible variation with time $t$, the schedule.

In second-order optimization, one can set $\eta$ proportional to the inverse Hessian of the loss function, but that is computationally infeasible for deep neural networks. 
And while quasi-Newton methods can approximate the inverse Hessian under certain conditions, in deep learning one usually adopts first-order methods with simple constant, oscillating, or decaying functions as schedules. 
The parameters of the schedule are determined by a hyperparameter search (often on a parameter grid), where the target architecture is trained repeatedly to determine a supposedly optimal $\eta_t^*$ schedule, which is then adopted for training of the production model or a model ensemble \citep{wu2018wngrad}.

But a fundamental understanding of what constitutes the optimal learning rate schedule for a specific model--data--task combination remains elusive \citep[e.g.][]{gotmare2018closer, li2019exponential, defazio2023optimal}.
It is therefore likely that neural networks could be trained more effectively and perform better if we understood what a good choice of $\eta_t$ needs to achieve. 

We will demonstrate that the training process undergoes distinct phases, whose occurrence can be recognized with a comprehensive view of the entirety of the training evolution. We present a meta-learning technique that models the network evolution as a dynamical system, described by a Latent Ordinary Differential Equation \citep[LODE;][]{chen2018neural, rubanova2019latent}, and then predicts the learning rate schedule $\eta(t)$ with the best long-term validation performance.
Our scheduler adapts to different network initializations and consistently outperforms any parametric schedule as well as more advanced gradient-dependent or RL schedules. It is computationally efficient, optimizer-agnostic, and can make use of experiment metrics tracked on platforms like Weights \& Biases or MLFlow \citep{Zaharia_Accelerating_the_Machine_2018, wandb}.

Our main contributions are:
\begin{itemize}
    \item We introduce an adaptive learning rate scheduler based on a latent ODE model to predict future training loss, validation performance, and learning rate trajectories.
    \item Our scheduler generates schedules that achieve superior test accuracy for CNN and ResNet models on the Fashion-MNIST, CIFAR-100, and ImageNet datasets and superior next-token prediction accuracy with a transformer model.
    \item Our method provides a practical and efficient way of navigating non-convex loss landscapes, we observe the tendency of the LODE scheduler to push learning rates beyond the ``edge of stability'' \citep{cohen2021gradient, arora2022understanding} and then settling in flatter minima than other schedules.
\end{itemize}

\section{Related work}

\paragraph{Parametric LR schedules.} 
The simplest explicit schedule adopts a constant learning rate, but has been shown to produce sub-optimal results \citep{smith2017don, gotmare2018closer, wu2019demystifying}.
Consequently, time-varying parametric learning rate schedules have been employed for decades \citep{darken1990note, darken1992learning}. Among the most successful of these schedules are cosine-decay, OneCycle \citep[often as cosine-OneCycle]{onecycle2017} and step-decaying schedules. Often these schedules are combined with other procedures such as warm-up to achieve their best results \citep{loshchilov2016sgdr}. A drawback of these methods is that they do not adapt to the training performance; instead, they choose the schedule shape and the corresponding hyperparameters either ad-hoc or through an extensive hyperparameter optimization. 

\vspace{-1em}
\paragraph{Schedule-free optimization.} \citet{defazio2024road} introduce an optimization framework that eliminates the need for predefined learning rate schedules and stopping time $T$. This is achieved through a novel form of momentum averaging. Their approach achieves strong empirical performance across vision, language, and recommendation tasks without requiring knowledge of the total training horizon or manual tuning of schedules, performing on-par or better than the baseline cosine and step-decay schedules. However, it still requires hyperparameter tuning and, because it does not depend on $T$, achieves almost all of its performance gains early during training.

\vspace{-1em}
\paragraph{Meta-optimizer.} \citet{baydin2017online} propose hypergradient descent, a method for dynamically adapting the learning rate during training by computing the gradient of the loss with respect to the learning rate itself. This method works for any gradient-descent optimization in a memory-efficient manner using the already computed values from the reverse-mode automatic differentiation. The method improves convergence speed and robustness across a range of tasks, particularly early in training, but remains unaware of the possible future performance. 

\vspace{-1em}
\paragraph{Reinforcement learning.} \citet{xu2019learning} and \citet{xiong2022learning} introduce RL-based approaches for adaptive learning rate scheduling, where a controller network is trained offline to adjust the learning rate based on features extracted from previous training runs, such as gradient statistics and validation loss. Rather than predicting the learning rate directly, the controller outputs a scaling factor applied to a pre-computed base schedule. While the method shows improved performance over fixed schedules on datasets like Fashion-MNIST and CIFAR-10, it requires significant upfront and per-iteration compute (see \autoref{sec:compute} for details). It also determines the current state from the parameter set of the trained network, which means that computational cost grows with the size of the network. And because rewards are based on changes of the loss and loss curves become increasingly shallow, late-time actions are only weakly constrained by training runs, so the method becomes unstable unless the adjustments of the learning rate are restricted to the narrow interval $[0.9, 1.1]$, foregoing more decisive changes to the learning rate even if those would be effective.

\vspace{-1em}
\paragraph{Learning curve extrapolation.} \citet{ding2025architecture} investigate the possibility of predicting late-stage performance of the optimization by extrapolating earlier training and test performance of MLP and CNN models. This is achieved through an initial encoding of the training trajectories via a graph-encoder, followed by modeling with a neural ODE to predict the dynamics. The authors use this method to rank hyperparameter choices while only observing a short initial optimization run. We demonstrate similar capability in \autoref{sec:lode-details} and extend the approach to generate new and improved schedules that have not been seen during prior optimization runs.

\vspace{-1em}
\paragraph{Sharp/dynamic LR transitions.} \citet{subramanian2024hop} study the effects of sharply changing precomputed learning rate schedules during the training of large transformer models on NLP tasks. They empirically find that such drastic changes can improve convergence and suggest there may be ways to leverage this behavior to produce more effective learning rate schedules. Our work will demonstrate an effective approach to introduce sharp changes for superior long-term results. 

\vspace{-1em}
\paragraph{Key distinctions.}  The main distinction of our LODE method is the ability to generate novel LR schedules from the summarized training metrics of past runs. While \citet{xu2019learning,xiong2022learning} also aim to improve scheduling via summarization of previous training information, their method applies a small multiplicative correction to existing schedules. Ding et al. (2025) use a graph-based ODE model to extrapolate training/validation metrics of parametric schedules with the goal of ranking them to find the most performant ones, but the chosen schedule itself is not altered. Both approaches therefore operate on a fixed, predetermined schedule, whereas our scheduler produces specialized schedules that deviate noticeably from any parametric form in its training data.

\section{Methods}
\vspace{-0.5em}
We present a scheduler that dynamically adjusts the current learning rate based on predictions of the long-term performance under the chosen schedule (see schematic in \autoref{fig:lode-schematic}). In what follows, we distinguish between the \emph{test model}, which we seek to optimize, and the \emph{LODE model}, which is used to learn the training behavior of the test model and choose the most effective schedule.

\subsection{LODE model}
\vspace{-0.5em}
We train a LODE encoder-decoder architecture \citep{chen2018neural, rubanova2019latent} to reconstruct the training behavior of the test model with metrics that are commonly logged during experiments like a hyperparameter search.  
In particular, the LODE model operates on three-dimensional time series $\mathbf{x}(t)\equiv(\ell(t), \nu(t), \eta(t))$, composed of the training loss $\ell$, a validation performance metric $\nu$, and the learning rate $\eta$ used by the schedule. Any suitable measure of performance computed on the validation dataset is acceptable as $\nu$; for image classification we will choose validation accuracy in our tests below, for NLP next-token prediction accuracy.
Note also that, unlike \citet{xu2019learning, xiong2022learning}, we do not include any information about the parameters of the test model. The training trajectory is thus a representation of the performance of the test network, not of its internal state. This choice allows our approach to be scaled to neural networks of arbitrary size without incurring extra costs during training or inference.

The LODE model first encodes the training time series of the test network into the latent vector
\begin{align}
\mathbf{z}_0 = \text{Encoder}_\phi\left(\left\{ \mathbf{x}(t_j)\right\}_{j \in \{i-\mu,\ldots, i\}} \right), 
\label{eq:latent_encoding}
\end{align}
where $t_i$ is the current step of the optimizer and $\mu$ the number of previous steps considered by the encoder. We choose an RNN architecture as an encoder similar to \citep{rubanova2019latent}. The latent vector $\mathbf{z}_0$ represents the joint training state, characterized by the network performance \emph{and} the learning rate used to achieve it. 

The latent vector is advanced to any time $t>t_i$ by integrating an ODE, represented by the neural network $f_\theta(\mathbf{z}, t)$:
\begin{equation}
\mathbf{z}(t) = \mathbf{z}_0 + \int_{t_i}^{t} f_\theta(\mathbf{z}(\tau), \tau) \, d\tau
\label{eq:int}
\end{equation}
During LODE training, we include an $\ell_2$ penalty in the loss function to minimize the path length of the latent trajectory, which encourages a time-invariant representation $\mathbf{z}$ \citep{auzina2024modulated, sampson2025path}. Its purpose is to ensure that states with similar training loss, learning rate, and validation metric are encoded to similar regions of latent space regardless of the time at which these values are observed.

From this latent state, one can generate the entire expected training behavior:
\begin{equation}
\mathbf{\hat x}(t) = (\hat{\ell}(t), \hat{\nu}(t), \hat{\eta}(t)) = \text{Decoder}_\psi(\mathbf{z}(t))
\label{eq:decoder}
\end{equation}
If the LODE model is accurate, we know how well the test model will perform when updated with a given optimizer and schedule at any point $t$ during its training.

\subsection{Dynamic Learning Rate Scheduler}
\label{sec:lode-scheduler}

We can now exploit our knowledge of the training process of the test network by changing the learning rate and predicting the resulting validation performance.
Our dynamic scheduler involves five steps to generate a learning rate at time $t_*$ during the optimization of the test network, which creates the test data $\mathbf{x}_*=(\ell_*, \nu_*, \eta_*)$.

\paragraph{1. Encoding} 
The time series segment $\mathbf{x}_*$ of length $\mu$ ending at $t_*$ is summarized by the LODE encoder into the latent vector $\mathbf{z}_0$ (\autoref{eq:latent_encoding}). For $t_* < \mu$, we initialize $\mathbf{z}_0$ with the data from the best-performing experiment in the LODE training data.
After every $\mu$ steps, the process is repeated to yield a new $\mathbf{z}_0$. Operating on such non-interleaving segments creates latents for optimization periods without our intervention, which more closely resemble the LODE training data.
\vspace{-0.5em}

\paragraph{2. Perturbation of the latent vector}
We create an ensemble of $n$ latent vectors $\mathbf{z}_{0,i} = \mathbf{z}_0 + \boldsymbol{\epsilon}_i$, where $\boldsymbol{\epsilon}_i\sim \mathcal{N}(0, \sigma^2)$. Each of these latents  is integrated via \autoref{eq:int} and decoded via \autoref{eq:decoder}, generating $n$ sample trajectories $\hat{\mathbf{x}}_i(t)$ of future training performance.
\vspace{-0.5em}

\paragraph{3. Sample acceptance} 
Of the potential trajectories, we select those samples $i$ that satisfy the following similarity condition: $\exists k\in[0, T]: |\hat{\ell}_i(t_k) - \ell_*(t_*)| < 2\,{\rm std}\left[\ell_*(t_j)\right]_{j \in \{*-\mu,\ldots, *\}}$, i.e. there must be at least one time $t_k$ during the optimization interval, for which the predicted loss is within 2 standard deviations of the recent losses of the test network. The requirement ensures that the selected sample is not just similar in latent space, but also close in loss to the test network at the current time. It permits the test network to reach that loss value at a different time than the prediction and accounts for the stochasticity of the training losses.
\vspace{-0.5em}

\paragraph{4. Goal conditioning}
Every accepted sample $i$ has at least one timestep $t_{k,i}$ with losses comparable to the current test network. We now compute the future validation metric $\hat\nu_{i,*}\equiv\hat{\nu}_i(t_{k,i} + \Delta t)$ we expect to achieve by following the future learning rate schedule $\hat{\eta}_i(t)$ from $t_{k,i}$ to $t_{k,i} + \Delta t$. The forward-looking horizon $\Delta t$ is a configuration parameter that controls how greedy the scheduling algorithm will be, from maximally greedy ($\Delta t=1$) to the longest period remaining in the optimization ($\Delta t = T-t_*$). If a sample $i$ has multiple times $t_k$ with similar losses to the test network, we choose the instance with the highest learning rate.
\vspace{-0.5em}

\paragraph{5. Ensemble predictor}
The final learning rate is computed for the next $\mu$ steps by averaging the three best-performing sample trajectories: 
$\tilde\eta(t_*+j)=\mathrm{mean}_{I_*}\left[\hat\eta_i(t_{k,i} + j)\right]$ for $j=0,\dots,\mu-1$, where $I_* = \mathrm{argsort}[\{\hat\nu_{i,*}\}]_{(0,1,2)}$.
The combination of trajectories improves the robustness of the predictor, especially in early phases of the optimization, where validation performance is often volatile.
Because of the rejection step 3, the number of similarly performing samples may be reduced to fewer than three, but those remaining will have the most appropriate learning rate predictions.
If no comparable sample exists, ideally because this scheduler outperforms any training run from the hyperparameter search, the current schedule will be extended for another $\mu$ steps. We perform an ablation study for the configuration parameters $(\mu, n, \sigma, \Delta t)$ of our scheduler in \autoref{sec:ablation}. 

We show pseudocode for the LODE scheduler in \autoref{alg:lode} below. This is also implemented in \texttt{lode$\_$scheduler.py} in the supplementary material.
\begin{algorithm}[t]
  \small
  \caption{Dynamic Learning Rate Scheduler}
  \label{alg:lode}
  \begin{algorithmic}[1]
      \\
      {\bf Requires} Current time step $t_*$, final time step $T$, trained LODE model $\mathcal{M}$,\\
      history of training loss, validation metric, learning rate 
      $\mathbf{X}_* = \{(\ell_*, \nu_*, \eta_*) (t_j)\}_{t_j=t_*-\mu}^{t_*}$;\\
      Configuration parameters: encoding length $\mu$, noise scale $\sigma$, ensemble size $n$, horizon $\Delta t$\\
    \Function{lode-scheduler}{$\mathbf{X}_*, t_*, T, \mathcal{M}$} 
            \State $\mathbf{z}_0 \gets \mathcal{M}.\mathrm{encode}(\mathbf{X}_*)$
            \Comment{1. Encode test data}
            \For{$i = 1$ {\bf to} $n$} \Comment{2. Create latent ensemble}
                \State {\bf if} $i=1$ {\bf then} $\boldsymbol{\epsilon}_i = \mathbf{0}$ {\bf else} $\boldsymbol{\epsilon}_i \sim \mathcal{N}(\mathbf{0},\sigma^2)$ \Comment{Keep direct latents in ensemble}
                \State $\mathbf{z}_{0,i} \gets \mathbf{z}_0 + \boldsymbol{\epsilon}_i$
                \State $\hat{\mathbf{X}}_i \gets \mathcal{M}.\mathrm{decode}(\mathbf{z}_{0,i};\ t=0, \dots, T+\Delta t)$
                \If {$\exists k\in[0, T]: |\hat{\mathbf{X}}_i.\ell(t_k) - \mathbf{X}_*.\ell(t_*)| < 2\;{\rm std}[\mathbf{X}_*.\ell]$}
                \Comment{3. Select similar loss}
                    \State $\hat\nu_{i,*}\gets\hat{\mathbf{X}}_i.\nu(t_{k,i} + \Delta t)$
                    \Comment{4. Estimate future performance}
                \EndIf   
            \EndFor
            \State $I_* \gets \mathrm{argsort}[\{\hat\nu_{i,*}\}]_{(0,1,2)}$
            \Comment{Select 3 best performers}
            \State $\tilde{\boldsymbol{\eta}}\gets\mathrm{mean}_{i\in I_*}[\hat{\mathbf{X}}_i.\eta(t_{k,i} + j)]$ {\bf for} $j=0,\dots,\mu-1$
            \Comment{5. Average over best samples}
        \State \Return $\tilde{\boldsymbol{\eta}}$ \Comment{Learning rate for the next $\mu$ steps}
    \EndFunction
  \end{algorithmic}
   \label{alg:lode}
\end{algorithm}

\section{Experimental setup}
\label{sec:baselines}
We compare our schedule to four non-adaptive learning rate schedules (constant, cosine-OneCycle, cosine-decay, step-decay), hypergradient descent \citep{baydin2017online}, the RL-based method from \cite{xu2019learning, xiong2022learning}, and the schedule-free optimization from \citet{defazio2024road}. We train a convolutional neural network (CNN), and ResNet18 \citep{he2016deep} on the classification tasks of the Fashion-MNIST and CIFAR-100 datasets. We also train a larger ResNet34 \citep{he2016deep} on the ILSVRC 2012 Imagenet dataset \citep{russakovsky2015imagenet}, as well as a small (28M parameters) transformer model \citep{radford2018gpt} tested on next-token prediction in a corpus of short stories \citep{eldan2023tinystories}. 
 We perform a hyperparameter sweep over a range of parameters for all LR schedulers, which is detailed in \autoref{sec:appendix-hypers}. For the Fashion-MNIST and CIFAR100 dataset-models we use the AdamW optimizer \citep{loshchilov2017decoupled}, for the ImageNet dataset we use stochastic gradient descent, and for the transformer model we use Adam \citep{kingma2014adam}. Once the best hyperparameters are found, we repeat training for multiple random initializations; all reported results are averaged over these initializations.

We train a LODE model to reconstruct the training loss, learning rate, and validation performance metric from a hyperparameter search comprising the four non-adaptive learning rate schedules. We omit schedules from hypergradient descent, RL-controller, and schedule-free methods in the training data to stay within the scope of conventional hyperparameter searches.
The specific architecture and training details of the latent ODE model are described further in \autoref{sec:lode-details}. We note that a separate LODE model is trained for each unique model/dataset pair.

\section{Results}
\label{sec:experiment-results}
\begin{wrapfigure}[31]{l}{0.5\linewidth}
    \vspace{-2em}
    \includegraphics[width=0.9\linewidth]{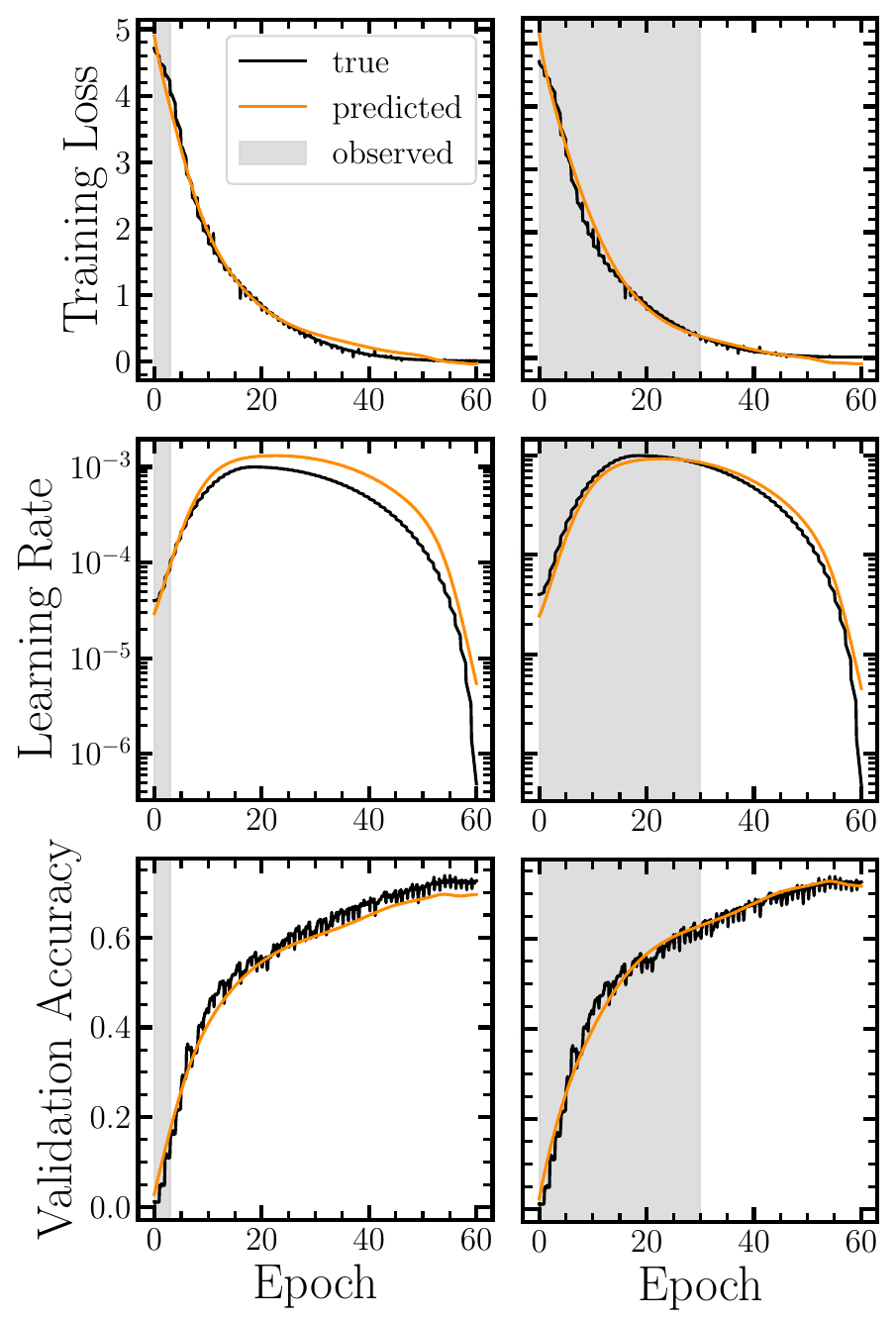}
    \caption{Training loss, learning rate, and validation accuracy observed for the CIFAR-100 trial (black) and prediction by the LODE model (orange) based on the first $5\%$ (left) and $50\%$ (right) of training epochs. Grey shading indicates the portion of the training trajectories encoded by the LODE model.}
    \label{fig:traj}
\end{wrapfigure}

\subsection{LODE modeling}
 \autoref{fig:traj} shows an example of the true (black) and reconstructed (orange) time series of training loss, learning rate, and validation accuracy for a single training run of the CIFAR-100 trial. We note that \textit{true} here means the actual learning rate, training loss, and validation accuracy recorded from a sample of the hyperparameter sweeps. The left column shows the results when observing only the first $5\%$ of the training periods (and predicting the remaining $95\%$), while the right column shows the results of observing the first $50\%$. In both cases, we see excellent reconstruction fidelity, which demonstrates that the training behavior, i.e. the relation between and time evolution of training loss, validation accuracy and learning rate, is recognizable even from very short segments of the trial run. We show further performance analysis of the LODE model in \autoref{sec:lode-details}.

\subsection{Optimizer performance}
\begin{figure}[h]
    \vspace{0em}
    \centering
    \includegraphics[width=0.98\linewidth]{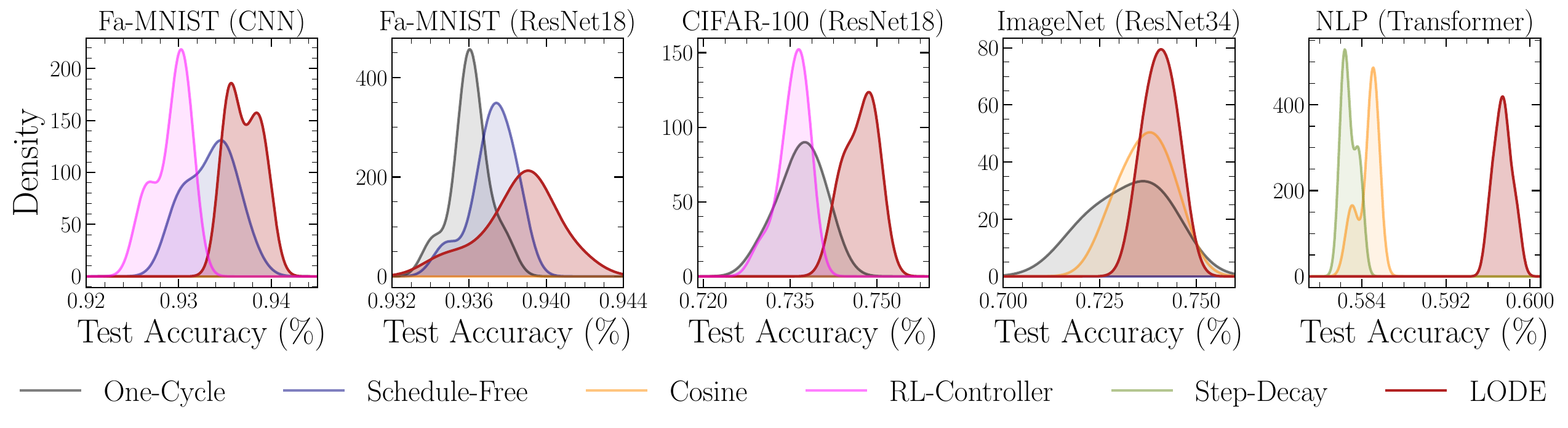}
    \vspace{-0.5em}
    \caption{Probability distribution of the final test accuracy for the test problems from \autoref{tab:results}. We run 20 trials for each schedule, varying the random seed for the network initializations, after choosing the best performing hyperparameters for each schedule. The test accuracy is measured at the epoch with the highest validation accuracy. Colors indicate the scheduling or optimization method; for clarity, we only show the three best performing methods in each panel.}
    \label{fig:kde}
    \vspace{-0.5em}
\end{figure}

We compare the final accuracy of the trained test model for the baseline schedules and our LODE scheduler in \autoref{tab:results}.
All results represent the best hyperparameter configuration of each schedule, with results averaged over 20 random seeds for Fashion-MNIST and CIFAR100 datasets, and 10 for ImageNet and TinyStories (more details in \autoref{sec:appendix-hypers}).
We also show the probability distribution of final accuracies for these tests in \autoref{fig:kde}. Our LODE scheduler outperforms all baselines for every model architecture and dataset. It achieves the highest-performing results very consistently, with small spread in the test accuracy.

\input{table1}

\subsection{Characteristics of the training path}
\begin{figure}[b]
    \centering
    \includegraphics[width=0.98\linewidth]{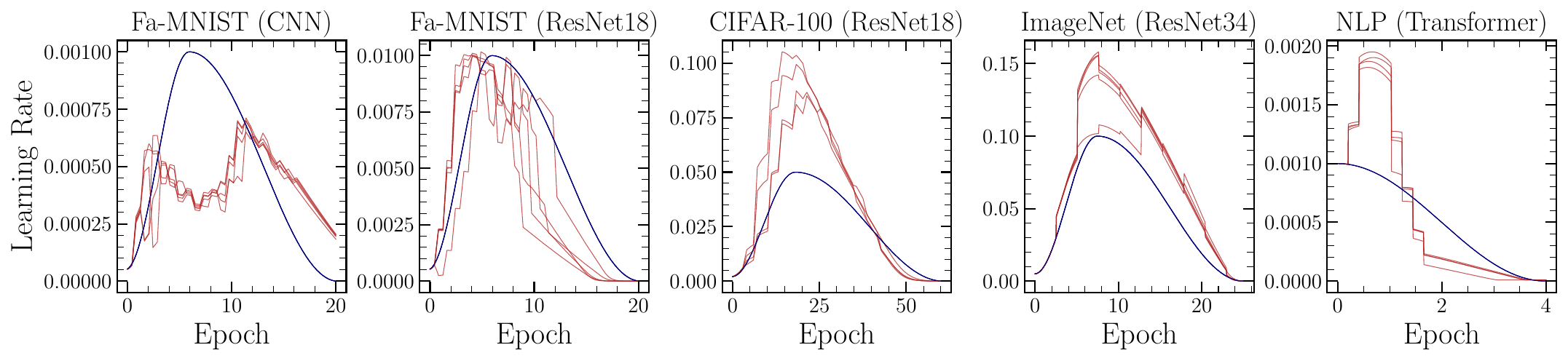}
    \vspace{-0.5em}
    \caption{Learning rate schedules generated by the LODE scheduler (red) for 5 random trials on the test problems from \autoref{tab:results}, compared to the best parametric schedule for each task (blue)}
    \label{fig:lrs}
\end{figure}

In \autoref{fig:lrs} we plot the learning rate sequences generated by the LODE scheduler for each model/dataset combination. It shows several noteworthy features:

\begin{itemize}
    \item The schedule shapes deviate noticeably from any parametric form in the training data.
    \item The change of either the model architecture on the same training data (CNN vs ResNet18 on Fashion MNIST) or of the training data for the same architecture (ResNet18 on Fashion MNIST vs CIFAR-100) leads to significant changes for high-performing schedules.
    \item Individual schedules show consistent patterns within each test case despite random initializations and the adaptability of the LODE scheduler.
\end{itemize}

To further explore the consequences of these new schedules, we compute the sharpness of the loss surface at convergence for the ResNet18 model on the CIFAR-100 dataset. Achieving wider, flatter minima has been found to indicate more generalizable and better performing models \citep{hochreiter1997flat, keskar2016large,jastrzkebski2017three, chaudhari2019entropy, petzka2021relative}.
Algorithms, such as stochastic weight averaging and sharpness-aware minimization \citep{izmailov2018averaging, foret2020sharpness} leverage this finding, even though a flatness condition alone should not be used as a success metric \citep{dinh2017sharp, kaur2023maximum}.

\begin{figure*}[h]
    \centering
    \includegraphics[width=0.98\linewidth]{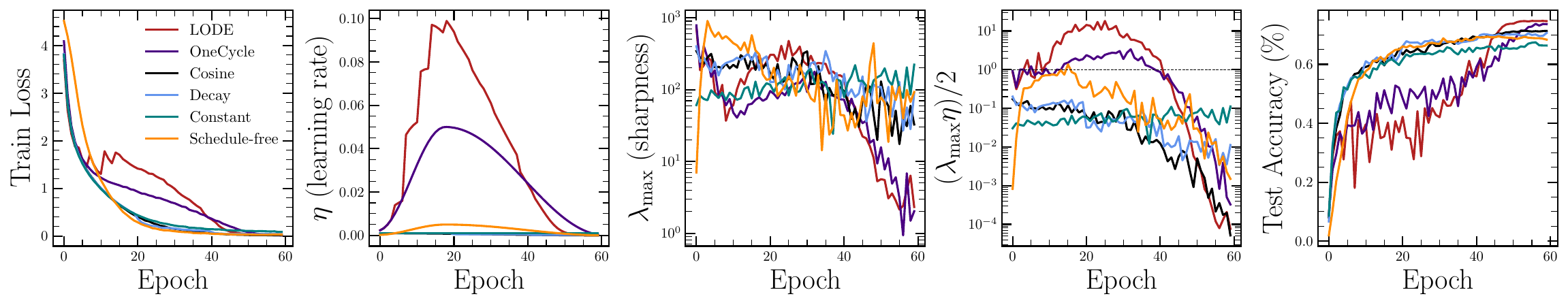}
    \vspace{-0.5em}
    \caption{\small Training loss (1st panel), Learning rate (2nd panel), loss sharpness (3rd panel), optimization stability (4th panel), and test accuracy (5th panel) for different schedules in the ResNet18--CIFAR-100 trial.}
    \label{fig:eos}
    \vspace{-0.5em}
\end{figure*}

We compute the sharpness of the minima achieved by each model by calculating the largest eigenvalue ($\lambda_{\rm{max}}$) of the approximate Hessian of the loss function with respect to the model parameters after optimization, using constant batch sizes of 256. The results are listed in \autoref{tab:eigvals} with an evolution over training shown in panel three of \autoref{fig:eos}. The LODE scheduler produces $\lambda_{\rm{max}}$ at least a factor two smaller than other schedulers evaluated at their best performance. The LODE scheduler therefore not only achieves superior final validation performance but should also improve model generalization.

\begin{table}[h]
    \centering
    \small
    \caption{Largest eigenvalue of the Hessian ($\lambda_{\rm{max}}$) for ResNet-18 on CIFAR-100. Computed via power-iteration; lower values indicate flatter minima.}
    \vspace{0.2em}
    \resizebox{0.85\textwidth}{!}{%
    \begin{tabular}{lccccccc}
    \toprule
    Schedule & LODE & OneCycle & Hypergrad & Schedule-free & Cosine & Decay & Const \\
    \midrule
    $\lambda_{\rm{max}}$ & 0.62$^{\pm0.22}$ & 1.22$^{\pm0.44}$ & 2.36$^{\pm0.37}$ & 2.66$^{\pm0.29}$ & 3.00$^{\pm0.29}$ & 3.69$^{\pm1.53}$ & 13.23$^{\pm3.56}$ \\
    \bottomrule
    \end{tabular}%
    }
    \label{tab:eigvals}
\end{table}

\subsection{Ablation Studies}
\label{sec:ablation}

\paragraph{Configuration parameters}
As described in \autoref{sec:lode-scheduler}, the LODE scheduler contains four configuration parameters: $n, \sigma, \mu, \Delta t$, representing the ensemble size, ensemble noise level in latent space, encoding length, and forward-looking horizon, respectively. We show the sensitivity of our scheduler to changes of the first two parameters with an ablation study performed with the CIFAR-100 dataset and the ResNet18 model in \autoref{tab:ablation}. We see that ensemble size and scatter have no significant effect on the scheduler performance and set defaults to $n=30, \sigma=0.15$.
The LODE performance test (\autoref{sec:lode-details}) indicates that $\mu$ as low as $5\%$ of the optimization epochs suffices for a high-fidelity reconstruction of the training dynamics, which we adopt as the default.

\begin{table}[h]
    \vspace{0em}
    \centering
    \small
    \caption{Validation accuracy averages over 5 runs on CIFAR-100 when varying LODE configuration parameters.}
    \label{tab:ablation}
    \vspace{0.0em}
    \resizebox{0.70\textwidth}{!}{%
    \begin{tabular}{lcccccc}
        \toprule
        $n$ & 10 & 10 & 30 & 30 & 60 & 60 \\ 
        $\sigma$ & 0.15 & 0.30 & 0.15 & 0.30 & 0.15 & 0.30 \\ \midrule
        Val Acc (\%) ↑ & 74.2$^{\pm1.2}$ & 73.7$^{\pm1.1}$ & 74.9$^{\pm0.9}$ & 73.6$^{\pm1.4}$ & 74.8$^{\pm1.2}$ & 73.9$^{\pm0.7}$ \\
        \bottomrule
    \end{tabular}%
    }
\end{table}

\begin{figure*}[h]
    \centering
    \includegraphics[width=0.99\linewidth]{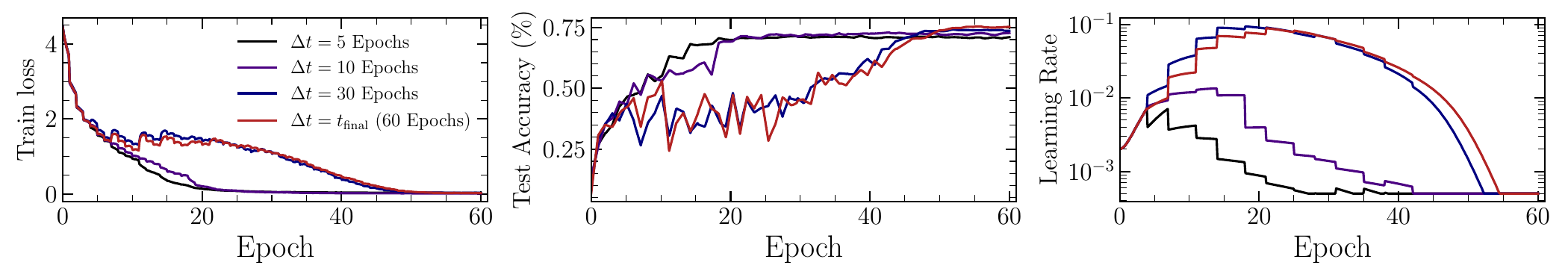}
    \vspace{-1em}
    \captionof{figure}{Training, test performance, and learning rate of the LODE-scheduler when changing the forward-looking horizon $\Delta t$ in the ResNet18--CIFAR-100 trial.}
    \label{fig:reward}
    \vspace{-0.5em}
\end{figure*}

\paragraph{Horizon}
With these defaults, we test the most important parameter, $\Delta t$, which determines how far in the future the estimated validation performance is evaluated (step 4 in \autoref{alg:lode}).
We vary the range from the minimum $\Delta t=\mu$, seeking to achieve the best performance at the end of the next training segment, to the maximum, $\Delta t=T - t_{\rm{current}}$, delaying optimal performance to the very end of the optimization.
The results are shown in \autoref{fig:reward}, again for training the ResNet18 model on CIFAR-100. 
It is evident that greedy adjustments are inferior to adopting a long-term goal.
A long horizon permits a longer period of exploration with drastically higher learning rates, which are reduced only shortly before the final epoch $T$ to permit a more gradual convergence. This long-term goal orientation is crucial for the superior performance of our scheduler and would be impossible to achieve without an accurate long-term prediction of the training dynamics by the LODE.

\paragraph{Latent ODE training data}
We also test the performance of the LODE scheduler as a function of the quantity and quality of the training data obtained during the hyperparameter search. This study is important to determine the upfront cost of training the LODE model or the possibility of simply exploiting existing experiment tracking results without any further generation of training data. We therefore assume a simple grid search with 4 parametric schedule shapes (constant, OneCycle, cosine-decay, step-decay) and 5 overall learning rates.
\autoref{tab:diversity} lists the most important results. Increasing the number of random network initializations with fixed hyperparameters mildly increases the LODE model's accuracy and thus the scheduler's efficacy, presumably because it illustrates the stochastic effects on the training. The largest impact stems from the presence of the OneCycle schedule in the training data: removing it leads to a significant drop in performance. But only using OneCycle during LODE training reproduces exactly the performance of the OneCycle schedule from \autoref{tab:results}.
We conclude that a modest number of experiments during the hyperparameter search will suffice to create a useful representation of the training process. 
\begin{table}[h]
    \centering
    \small
    \caption{Test accuracy in CIFAR-100 when varying data volume (number of random initializations) and schedule diversity to train the LODE model.}
    \vspace{0.2em}
    \resizebox{0.7\textwidth}{!}{%
    \begin{tabular}{lccccc}
    \toprule
    Setting & 1 init & 5 inits & 10 inits & no OneCycle & only OneCycle \\
    \midrule
    Test Acc (\%) $\uparrow$ & 73.9$^{\pm0.9}$ & 74.9$^{\pm0.9}$ & 74.8$^{\pm0.9}$ & 72.4$^{\pm1.5}$ & 74.0$^{\pm1.1}$ \\ 
    \bottomrule
    \end{tabular}%
    }
    \label{tab:diversity}
    \vspace{-1em}
\end{table}


\vspace{-0.5em}
\subsection{Computational cost}
\label{sec:compute}
\vspace{-0.5em}

\begin{table*}[h]
\centering
\caption{Walltime per epoch for different schedules and optimization techniques in the ResNet18--CIFAR-100 trial on a single NVIDIA-A100. Runtimes are averaged over 5 trials of 20 epochs. The update frequency $\mu$ is reported in percentage of total epochs, $n$ denotes the latent ensemble size.}
\vspace{-0.5em}
\label{tab:runtime_ablation}
\resizebox{0.95\textwidth}{!}{%
\begin{tabular}{cccccccccccc}
    \toprule
     & \multicolumn{6}{c}{LODE Scheduler} & \multicolumn{1}{c}{Parametric} & \multicolumn{1}{c}{Hypergrad} & \multicolumn{1}{c}{Schedule-Free} & RL-Schedule \\ \cmidrule(lr){2-7} \cmidrule(lr){8-8} \cmidrule(lr){9-9} \cmidrule(lr){10-10} \cmidrule(lr){11-11}
    $\mu$       & 10\% & 5\% & 2.5\% & 10\% & 5\% & 2.5\% & -- & -- & -- & -- \\
    $n$         & 20  & 20 & 20 & 40 & 40 & 40 & -- & -- & --& -- \\
    Time (s)    &5.11$^{\pm .1}$ & 5.47$^{\pm.1}$ & 6.09$^{\pm.1}$ & 5.98$^{\pm .1}$ & 6.12$^{\pm .1}$ & 6.28$^{\pm.1}$ & 4.79$^{\pm .0}$ & 11.68$^{\pm .2}$ & 5.01$^{\pm .1}$ & 11.2$^{\pm .4}$ \\
    \bottomrule
\end{tabular}%
}
\end{table*}

The LODE scheduler incurs two additional costs compared to parametric schedules: the initial training of the LODE model, and, at test time, the integration of training dynamics for each member of the latent ensemble. 
\autoref{tab:runtime_ablation} lists the test-time cost per epoch for the LODE-scheduler with a range of hyperparameter configurations compared to the other schedules and optimization methods. With default parameters, the LODE schedule is about $25\%$ more expensive than simple parametric schedules, but significantly cheaper than hypergrad optimization and the RL scheduler. It is also important to note that the inference cost of the LODE does not increase with model size (unlike RL methods).

As with any meta-learning method, the cost of training the LODE model is dominated by the cost to create the training data from the test network. In general, we find using 5 random seeds of 4 parametric schedules, swept over 5 different learning rate values sufficient across all domains. This represents an upfront cost of 100 training runs. The exact cost for the LODE pipeline can be estimated as roughly 100 times the cost of a single parametric trial, assuming one uses the same suite of parametric schedules, with 5 seeds each. The training of the LODE model itself, which requires no hyperparameter tuning (see \autoref{sec:lode-details}), took less than one hour on a single NVIDIA A100. For comparison, the RL controller required two hours of offline training for the ResNet18 model. 
The computational cost of running dozens of trial optimizations is also incurred with other hyperparameter tuning methods or simple grid searches. But by utilizing the entire training evolution, instead of merely its final state, our scheduler produces better performing and generalizing results than those alternatives.  Methods for reducing the training cost will be investigate in the future.

\section{Discussion}

By learning from the training behavior a test network exhibits during a conventional hyperparameter grid search, we can 1) determine an effective representation of the current training state, 2) predict with high accuracy the future performance when following a proposed schedule, and 3) create a specialized schedule to achieve the best predicted long-term validation performance under a given optimization objective. Following this schedule indeed leads to minimizers with superior validation results, located in smooth regions of the loss function.

This outcome is surprising for two reasons. Any dynamical model requires a suitable state representation, but our approach constructs this state representation from three simple quantities that are routinely tracked in ML experiments. Unlike the RL-scheduler of \citet{xu2019learning, xiong2022learning}, information about the parameters of the network is not needed. The LODE state therefore represents the \emph{effective} training behavior, which is legitimate because in deep learning architectures all minima are practically equivalent \citep{choromanska2015loss, ge2016matrix}. Our approach can be applied to test networks of arbitrary size because it does not seek to learn the dynamics of the network parameters, only of the aggregate network performance.
The latent representations thus indicate training \emph{phases}: e.g. the initial phase of gathering a reliable gradient direction; an exploration phase with high volatility; a convergence phase where the loss becomes shallow; and possibly an overfitting phase with divergence of training and validation performance  (see \autoref{fig:umap} for a visual confirmation of these training phases). Recognizing these phases allows the scheduler to adapt the learning rate such that it achieves high validation performance at the end of the optimization period.

The second surprise lies in the stability of the optimization despite the large learning rates that are, at certain times, proposed by the LODE scheduler. This empirical finding is consistent with works by \citet{cohen2021gradient, arora2022understanding}, who also find that learning rates for deep neural networks operate at and often substantially exceed the limits of traditional stability: For quadratic loss functions, gradient updates remain stable only for learning rates $\eta<2/\lambda_{\rm max}$, where  $\lambda_{\rm max}$ is the largest eigenvalue of the Hessian of the loss function. 
\autoref{fig:eos} shows the relevant quantities for all schedules in the ResNet18--CIFAR-100 trial. The horizontal line in the fourth panel indicates the traditional ``edge of stability'' (EoS) criterion. Several schedules track this line initially, but the LODE scheduler and, to a lesser degree, the OneCycle schedule exceed the EoS limit in the middle phase of the optimization. Not only do both remain stable, but their investment in a longer exploration phase yields better long-term results than those schedules with smaller $\eta$.

The third panel of \autoref{fig:eos} shows the time evolution of the sharpness of the loss, revealing a noticeable reduction after $t\approx 40$, indicating increasing model generalization at the same time as when the learning rate is decreased and the validation accuracy makes the most important late-time gains. 
While earlier works (\citealt{jastrzkebski2017three}, \citeyear{jastrzkebski2018relation, jastrzebski2020break}) have demonstrated that large learning rates lead SGD into areas of the loss function with lower sharpness, the picture that now emerges for network training is more nuanced.
Because loss landscapes for neural networks exhibit multi-scale properties \citep{ma2022}, it is favorable to traverse the loss landscape with a sufficiently high learning rate to find the largest, flattest basin nearby. Not only is the interior of such a basin less rough, the number of wide and well-connected minima it contains scales with a power-law of its size \citep{ly2025Nature}. 
But traversing the loss landscape at that speed leads to highly non-monotonic losses, for which only a long-term average is guaranteed to decrease \citep{arora2022understanding}.
To ensure good final performance for a finite-length optimization, learning rates must decline to reduce the stochastic component of the loss.
The LODE scheduler recognizes these different phases and predicts when and how the transition from the exploration to convergence phase needs to occur.
We surmise that an optimizer with our schedules is better able to find and settle in large basins, which explains the significantly reduced sharpness in \autoref{tab:eigvals}. This ability appears to become more important in higher dimensions, where the basins occupy a smaller fraction of the volume, as shown by the largest improvements in \autoref{fig:kde} being attained for the largest model, the transformer.

\vspace{-0.5em}
\section{Conclusion}
We present a generative learning rate scheduler, which uses metrics commonly gathered during hyperparameter searches to create a dynamical systems representation of the neural network training process.
By leveraging the entire time evolution of the optimization under different schedules---instead of merely the final outcome---our method assembles an optimal schedule for a given model--task--dataset combination.
It predicts which schedule will lead to the best long-term performance and achieves superior test accuracy and model generalization across image classification and text generation tasks.
The significance of the gains from employing the LODE scheduler become more pronounced for more complex loss landscapes, with the largest improvement observed for the transformer, our largest model.
Our method is computationally efficient, numerically stable, and creates new, specialized paths to superior results in high-dimensional gradient descent.

\subsubsection*{Acknowledgments}
This work was supported by a grant from the W. M. Keck Foundation.


\bibliography{refs}
\bibliographystyle{tmlr}

\appendix
\onecolumn
\setcounter{table}{0}
\renewcommand{\thetable}{A\arabic{table}}
\renewcommand*{\theHtable}{\thetable}

\setcounter{figure}{0}
\renewcommand{\thefigure}{A\arabic{figure}}
\renewcommand*{\theHfigure}{\thefigure}

\setcounter{algorithm}{0}
\renewcommand{\thealgorithm}{A\arabic{algorithm}}

\section{Appendix - Experimental details}

\subsection{Model architectures}

\paragraph{CNN}
The CNN is designed for Fashion-MNIST dataset. We apply two consecutive $3\times3$ convolution–ReLU layers (64 channels) followed by max pooling and dropout ($p=0.25$), then repeat the pattern with 128-channel convolutions. The flattened features pass through a 256-unit fully connected layer with ReLU and dropout before a final dense layer.

\paragraph{ResNet}
We follow the implementation from \citep{he2016deep} creating an 18-layer ResNet for use on the Fashion-MNIST and CIFAR-100 datasets, and a 34-layer ResNet for the ImageNet dataset.

\paragraph{Transformer (miniGPT)}
We train a miniGPT model \citep{radford2018gpt} for our NLP task. Each Transformer block applies masked multi-head self-attention  followed by a feed-forward MLP with ReLU activations, with dropout ($p=0.1$) and residual connections around each sub-layer. Both sub-layers use residual connections with post-layer-normalization ($\epsilon$ = 1e-6) and dropout ($p = 0.1$). We use 256-dimensional token embeddings combined with learned absolute positional embeddings, a feed-forward width of 256, and a context length of 256 tokens. Text is tokenized with the GPT-2 byte-pair encoding (vocabulary 50,257). We train on the TinyStories corpus \citep{eldan2023tinystories}, with separate validation and test splits used for the reported next-token prediction accuracy. We sweep over the learning rates as shown in \autoref{tab:results}.

\subsection{Baseline details}
\label{sec:appendix-hypers}
We compare against parametric, hypergradient, schedule-free and RL baselines. We perform hyperparameter sweeps for each baseline with all results presented in the main paper coming from the best performing models.  
\paragraph{Parametric models}
We use the built-in learning rate schedules from \textsc{optax} for cosine-decay, cosine-OneCycle, step-decay and constant learning rate schedules. For the Fa-MNIST and CIFAR-100 trials we perform sweeps over the learning rate (initial for cosine-decay and step-decay, peak-lr for OneCycle) for $\eta \in \{10^{-3},\ 5 \times 10^{-3},\ 10^{-2},\ 5 \times 10^{-2},\ 10^{-1}\}$. We use \textsc{AdamW} for all trials and try both with and without Nesterov momentum finding no significant differences. For the CNN models on the Fa-MNIST data we use weight decay of $10^{-3}$, for the ResNet18 on Fa-MNIST we use weight decay of $10^{-4}$, and for the ResNet18 on CIFAR-100 we use weight decay of $10^{-2}$ for all trials. 

For the ResNet34 on the ImageNet dataset we use standard SGD with Nesterov momentum in all trials and a weight decay of $10^{-4}$. We sweep over higher learning rates than the other experiments with $\eta \in \{0.005,\ 0.01,\ 0.05,\ 0.1,\ 0.256\}$. 

Once we find the optimal hyperparameters for each parametric model we perform 10 additional trials varying the random seed for both weight initializations, and training data shuffling.

\paragraph{Hypergradient descent}
We implement hypergradient descent as in \citep{baydin2017online} to calculate the individual learning rate updates. We perform the same $\eta$ sampling as in the above parametric models for the Fa-MNIST, CIFAR-100 and ImageNet datasets where $\eta$ represents the initial flat value of the learning rate used.

\paragraph{Schedule-free}
We use the schedule-free learning rate implementation directly from the \textsc{optax} library, ensuring to set the underlying $\beta_1$ values in the AdamW optimizer initially to 0, resetting them to the optimal value we found ($0.9$) in the instantiation of the schedule-free optimizer. We perform the same $\eta$ sampling as the above for all datasets. We make sure to use the correct evaluation parameters as explained in the documentation via running 
\texttt{eval$\_$params = optax.contrib.schedule$\_$free$\_$eval$\_$params(opt$\_$state, params)} 
before assessing validation, and test performance.


\paragraph{RL-controller}
We implement the RL-controller explained in \citep{xu2019learning} and detailed here \href{https://github.com/nicklashansen/adaptive-learning-rate-schedule}{https://github.com/nicklashansen/adaptive-learning-rate-schedule}. We made minor changes to resolve version conflicts with outdated Python packages. We use a version of the RL-controller provided in the supplementary material of \citep{xiong2022learning} for the graph-based RL controller. We again perform the same hyperparameter sweeps as with the other baselines. We note that, due to extremely long training times, we do not train this model on the ResNet34, ImageNet, or transformer model.

\subsection{LODE architecture and training}
\label{sec:lode-details}

\begin{figure}[h]
    \centering
    \includegraphics[width=0.98\linewidth]{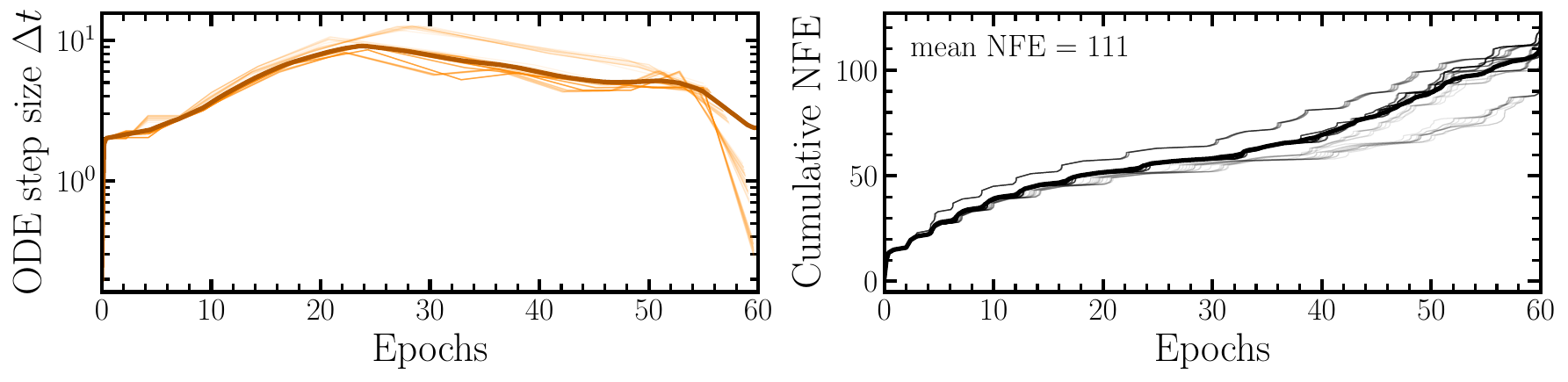}
    \caption{Adaptive ODE‑solver step size (left) and cumulative function evaluations (right) versus epoch for the latent‑ODE integration step of 100 random CIFAR‑100 runs (Tsit5, rtol $=$ atol $= 10^{-5}$). Faint lines show individual runs with bold showing the mean ($\sim$111 evaluations per full trajectory).}
    \label{fig:stiffness}
\end{figure}

We train a latent ODE (LODE) model for each dataset--model combination, resulting in 5 final trained LODEs. The training data for each LODE consists of the training trajectories of the parametric schedulers. We use 5 random trials for each schedule, resulting in 100 training trajectories for each LODE model. For simplicity, and to demonstrate efficient use of this method without any hyperparameter tuning of the LODE model itself, we use identical architectures for all trained LODEs with a single training run per model. We use latent and hidden dimensions of 20, and the ODE function is modeled with an MLP with 2 layers of 20 units with Tanh activations. We use an ODE-RNN encoder as in \citet{rubanova2019latent} however, we remove the variational penalty as in \citep{sampson2025path}. We train with a batch size of 20 and use a OneCycle schedule with a peak learning rate of $0.001$ using \textsc{Adam}. The maximum training time of the LODE models used 50,000 gradient updates, taking just under 1 hour on a single NVIDIA A100 GPU. All 5 models converged below 1 hour of GPU time, and we performed no hyperparameter tuning. We use \texttt{diffrax} \citep{kidger2021on} with a Tsitouras' 5(4) method (\texttt{Tsit5}). Integration is adaptive with step sizes controlled by a PID controller with relative and absolute tolerances both set to \texttt{rtol} = \texttt{atol} = $1\rm{e}-5$, an initial step size of $dt = 0.1$, and all computation performed in 64-bit precision. These settings were fixed across all five dataset–model LODEs and again were not tuned. \autoref{fig:stiffness} shows a plot of the ODE integrator's adaptive step size, and cumulative function evaluations to provide insight into the potential stiffness of the ODEs in latent space. We see stable step size evolution, with no spiking, which is consistent across 100 randomly sampled trials.

\begin{figure}
    \centering
    \includegraphics[width=0.99\linewidth]{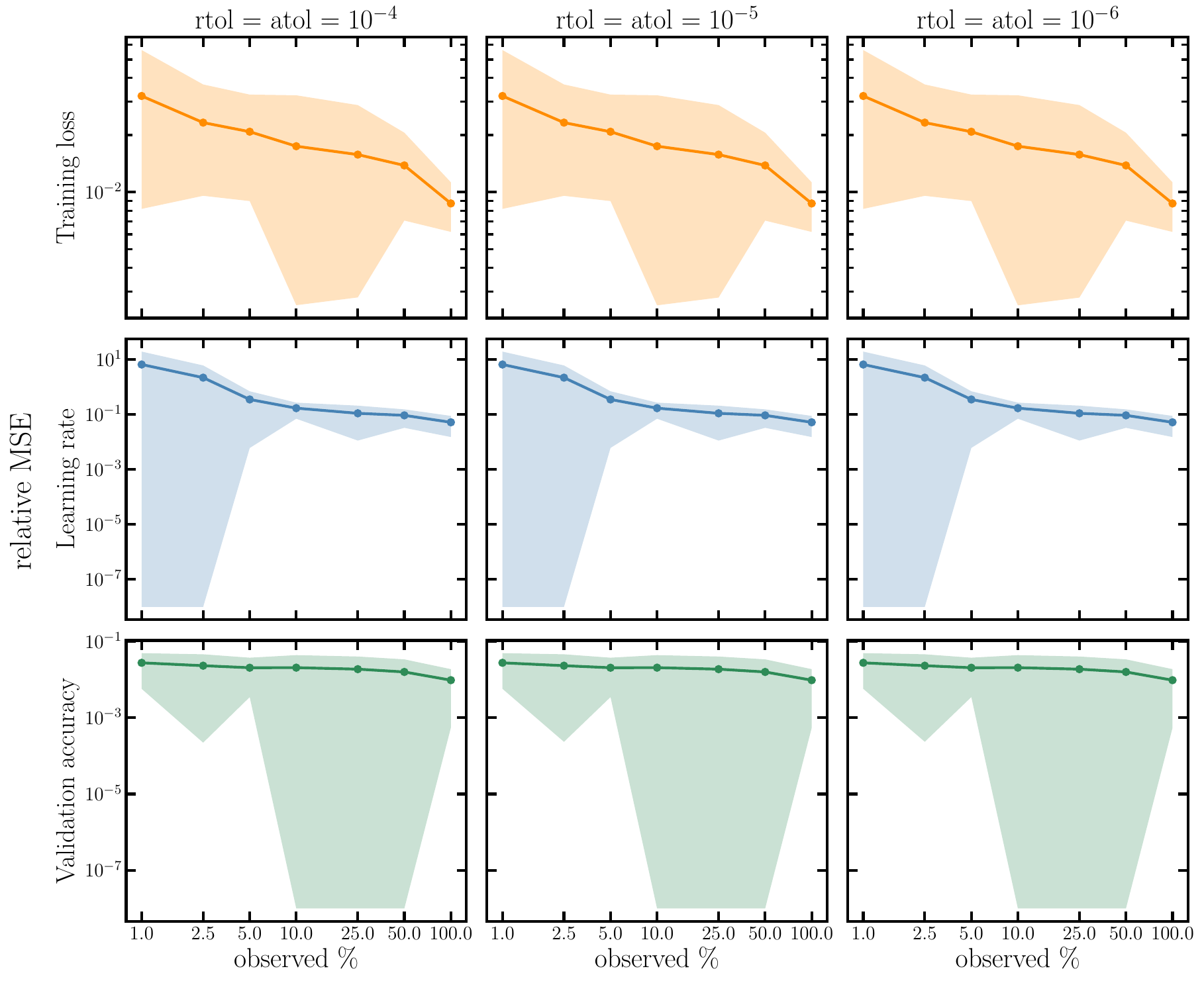}
    \caption{Relative MSE between the latent‑ODE reconstruction and the ground‑truth ImageNet ResNet‑34 training curves, as a function of the percentage of each run shown to the model (log–log axes). Rows are the three predicted quantities (training loss, learning rate, validation accuracy) with the columns showing the ODE solver tolerance (rtol $=$ atol $= 10^{-4}, 10^{-5}, 10^{-6}$). Solid lines and shaded bands show the mean and $\pm1\sigma$ across all averaged schedules, learning rates, and seeds.}
    \label{fig:mse_ablation}
\end{figure}

To determine the efficacy of our approach it is important to validate the reconstruction fidelity of the trained LODE model. The reconstruction accuracy of the trained LODE models for each trial is shown in \autoref{tab:predictions}. We see very accurate predictions even when using as little as $5\%$ of the data to extrapolate the full training trajectories. We show an ablation of the relative MSE from the LODE generation over the integrator error tolerances, and observation fraction in \autoref{fig:mse_ablation}.

\begin{table*}[h]
    \small 
    \centering
    \caption{Relative MSE for latent ODE predictions averaged across 5 random seeds per schedule for all dataset--model tests when observing either the initial $5\%$ or the initial $50\%$ of the training time series.}
    \vspace{0.5em}
    \resizebox{0.98\textwidth}{!}{%
    \begin{tabular}{cccccc}
    \toprule
        \textbf{Metric} & Fa-MNIST (CNN) & Fa-MNIST (ResNet18) & CIFAR-100 (ResNet18) & ImageNet (ResNet34) & NLP (Transformer)\\ \midrule 
        train loss ($5\%$) & 0.177$^{\pm.041}$ & 0.105$^{\pm.025}$ & 0.028$^{\pm.006}$ & 0.021$^{\pm.003}$ & 0.036$^{\pm.011}$\\ \rowcolor{gray!15}
        train loss ($50\%$) & 0.119$^{\pm.028}$ & 0.106$^{\pm.025}$ & 0.024$^{\pm.006}$ & 0.014$^{\pm.002}$ & 0.008$^{\pm.002}$\\
        learning rate ($5\%$)& 0.287$^{\pm.074}$ & 0.340$^{\pm.081}$ & 0.335$^{\pm.078}$ & 0.349$^{\pm.051}$ & 0.231$^{\pm.052}$\\ \rowcolor{gray!15}
        learning rate ($50\%$) & 0.241$^{\pm.057}$ & 0.151$^{\pm.081}$ & 0.132$^{\pm.078}$ & 0.100$^{\pm.011}$& 0.131$^{\pm.050}$ \\
        val accuracy ($5\%$) & 0.041$^{\pm.007}$ & 0.031$^{\pm.001}$ & 0.011$^{\pm.003}$ & 0.020$^{\pm.003}$ & 0.051$^{\pm.013}$\\ \rowcolor{gray!15}
        val accuracy ($50\%$) & 0.010$^{\pm.002}$ & 0.003$^{\pm.001}$ & 0.011$^{\pm.003}$ & 0.016$^{\pm.002}$ & 0.021$^{\pm.003}$\\
         \bottomrule
    \end{tabular}
    }
    \label{tab:predictions}
\end{table*}

\paragraph{Ability to select optimal trials}
We also determine if our model is able to identify the optimal performing learning rate schedules from the data. To test this we supply the latent ODE with the first epoch of training data for each parametric schedule configuration and then ask the model to extrapolate which trial will have the largest validation accuracy at the final epoch. We perform this test for each dataset--model combination, listing the ranking of the final validation accuracy of the chosen trial (out of 100 trials) with the final validation accuracy of that trial, as well as the final accuracy of the single best trial. As we are not averaging over seeds, the maximum final accuracy reported here is slightly higher than in \autoref{tab:results}. In all tests, we correctly ranked the best performing hyperparameters. The results below show our ability to adequately distinguish between best performing individual seeds within each hyperparameter trial.
\begin{itemize}[noitemsep, topsep=0pt, parsep=1pt, partopsep=0pt]
    \item FaMNIST (CNN): selected $2^{nd}$ best trial; final accuracy $94.3\%$; best final accuracy $94.3\%$
    \item FaMNIST (ResNet): selected $10^{th}$ best trial; final accuracy $93.9\%$; best final accuracy $94.3\%$
    \item CIFAR100: selected $4^{th}$ best trial; final accuracy $73.8\%$; best final accuracy $74.1\%$
    \item ImageNet: selected $2^{nd}$ best trial; final accuracy $74.4\%$; best final accuracy $74.7\%$
    \item Transformer: selected $2^{nd}$ best trial; final accuracy $57.46\%$; best final accuracy $57.47\%$
\end{itemize}
We see that via observing the training metrics of first epoch, our trained LODE models perform very well in determining which trial leads to the best final validation accuracy. This result, combined with the accurate reconstruction of the learning rate schedules demonstrated in \autoref{fig:traj} and \autoref{tab:predictions}, give us confidence that the trained LODE models are learning useful latent representations that capture the underlying training dynamics and can reliably guide learning rate selection from early signals.

\subsection{Choice of evaluation metrics}
To determine the \textit{optimal} trajectory we choose validation accuracy rather than validation loss for two reasons. 1) In general, the validation accuracy provides a smoother curve, making LODE training and inference easier. 2) In most cases one is more interested in the final accuracy (or accuracy-related metrics such as the F1 score) than the loss. As the gradient updates are already guided entirely by minimizing the loss, including information about validation accuracy in the optimization routine is preferable.

\subsection{Latent representations}
\begin{figure*}
    \centering
    \includegraphics[width=0.98\linewidth]{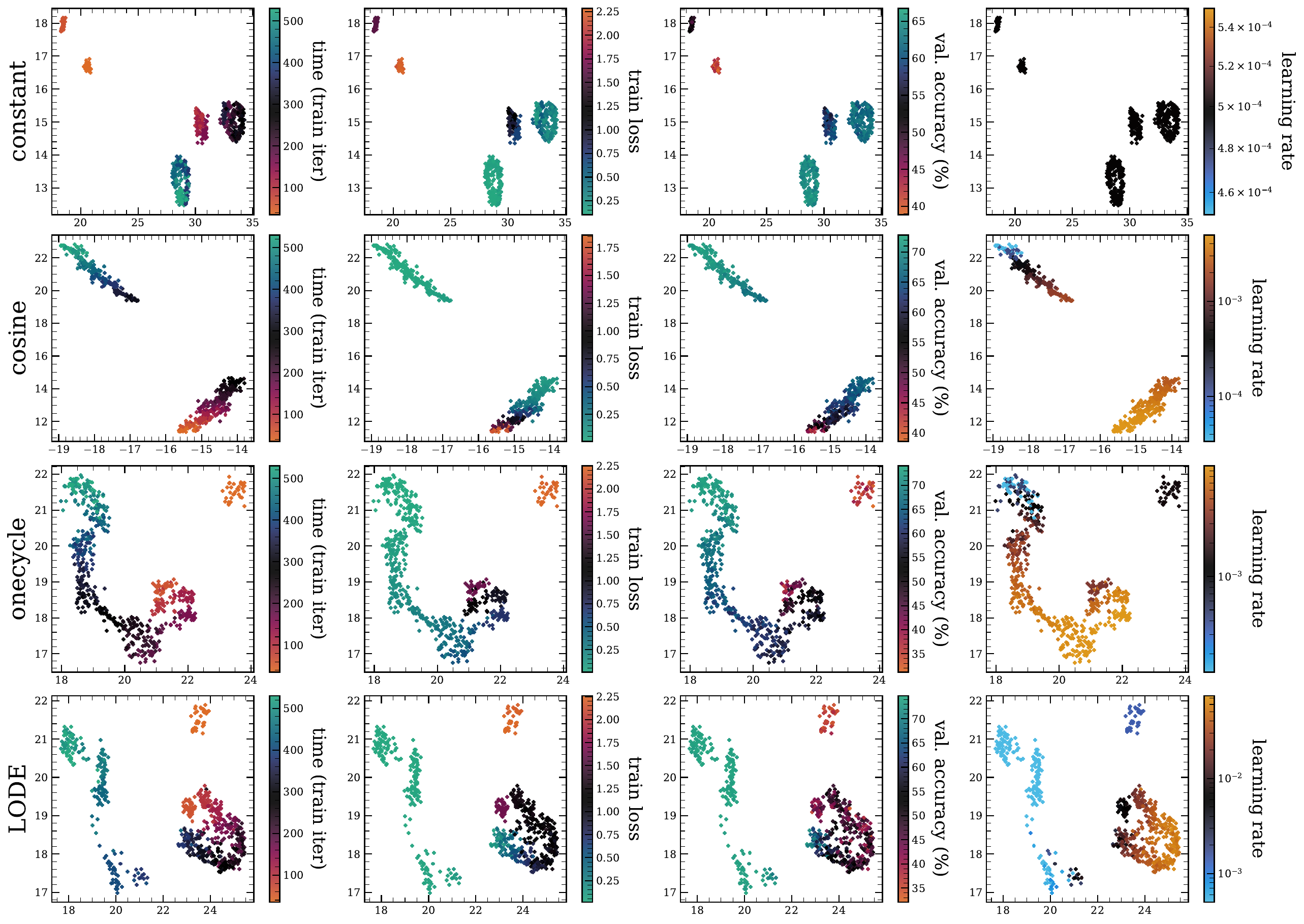}
    \caption{UMAP of the latent vectors generated by the LODE-encoder for a series of parametric, and LODE-guided trials of training on the ResNet18--CIFAR-100 trial. We show the training time, train loss, validation accuracy, and learning rate.}
    \label{fig:umap}
\end{figure*}

In \autoref{fig:umap} we show a UMAP \citep{mcinnes2018umap} of the latent encodings created by our LODE encoder for 20 random trials of training a ResNet18 on CIFAR-100. We show results for 3 parametric schedules of constant, cosine-decay and onecycle, as well as the embedding from trials using the LODE, and show 10 points per epoch per trial. We note stark differences between the latent embeddings for the constant, and cosine schedules compared to the onecycle and LODE schedule. We can also see that the LODE model finds distinct clusters corresponding to the initial optimization phase, the exploration phase, and the convergence phase. These representations allow the LODE scheduler to shape the optimization in such a way to achieve the best long-term performance.

\end{document}

%% file: math_commands.tex
\usepackage{amsmath,amsfonts,bm}

\def\eqref#1{equation~\ref{#1}}

\def\1{\bm{1}}

\DeclareMathAlphabet{\mathsfit}{\encodingdefault}{\sfdefault}{m}{sl}
\SetMathAlphabet{\mathsfit}{bold}{\encodingdefault}{\sfdefault}{bx}{n}



%% file: table1.tex
 \begin{table}[t]
    \vspace{1em}
    \centering
    \caption{Comparison of final test accuracy for all  schedules, models, and datasets (details in \autoref{sec:baselines}).
    We report the mean $\pm$ standard deviation for 20 random seeds (10 for ImageNet). For the NLP task the test accuracy is next-token prediction accuracy.
    Our LODE-scheduler achieves superior results for every model/dataset pair. The RL-controller was not evaluated by us on ImageNet or the NLP task due to high computational cost. The ImageNet values listed here are taken from \citet{xiong2022learning}. Note, in the bottom row, we report the $p$-value from a two-sided t-test between the LODE and the next most performant scheduler.}
    \label{tab:results}
    \vspace{0.5em}
    \setlength{\tabcolsep}{3pt}
    \resizebox{0.95\textwidth}{!}{  
    \begin{tabular}{lccccccc}
    \toprule
      & \multicolumn{2}{c}{\textbf{Fashion MNIST}} & \textbf{CIFAR-100} & \textbf{ImageNet} & \textbf{NLP} \\
    \cmidrule(lr){2-3} \cmidrule(lr){4-4} \cmidrule(lr){5-5} \cmidrule(lr){6-6}
    & CNN & ResNet18 & ResNet18 & ResNet34 & Transformer \\
    \textbf{Scheduler} & Test Acc (\%) ↑ & Test Acc (\%) ↑ & Test Acc (\%) ↑ & Test Acc (\%) ↑ & Test Acc (\%) ↑ \\
    \midrule
    Constant       & 91.5$^{\pm0.1}$ & 91.9$^{\pm0.5}$ & 67.3$^{\pm0.6}$ & 61.7$^{\pm0.4}$ & 57.9$^{\pm0.1}$ \\
    Cosine         & 92.5$^{\pm0.1}$ & 93.5$^{\pm0.1}$ & 71.2$^{\pm1.4}$ & 73.9$^{\pm0.4}$ & 58.5$^{\pm0.1}$ \\
    OneCycle       & 92.9$^{\pm0.2}$ & 93.6$^{\pm0.1}$ & 74.0$^{\pm1.2}$ & 73.8$^{\pm0.5}$ & 58.1$^{\pm0.2}$ \\
    Step Decay          & 92.4$^{\pm0.2}$ & 93.5$^{\pm0.1}$ & 69.6$^{\pm0.9}$ & 66.1$^{\pm0.5}$ & 58.3$^{\pm0.1}$ \\
    Hypergrad      & 92.8$^{\pm0.1}$ & 91.7$^{\pm0.5}$ & 70.3$^{\pm1.0}$ & 59.9$^{\pm0.5}$ & 55.3$^{\pm0.4}$ \\
    RL-controller  & 93.0$^{\pm0.1}$ & 92.9$^{\pm0.4}$ & 73.9$^{\pm0.9}$ & 71.6$^{\pm1.0}$ & -- \\
    Schedule-free  & 93.6$^{\pm0.2}$ & 93.7$^{\pm0.1}$ & 71.1$^{\pm1.0}$ & 60.2$^{\pm0.8}$ & 56.3$^{\pm0.2}$ \\
    \rowcolor{gray!15}
    LODE (ours)    & \textbf{93.8$^{\pm0.2}$} & \textbf{93.9$^{\pm0.3}$} & \textbf{74.9$^{\pm0.9}$} & \textbf{74.5$^{\pm0.2}$} & \textbf{59.8$^{\pm0.2}$} \\
    $p$-value & 0.003 & 0.01 & 0.01 & 0.0005 & 0.0001 \\
    \bottomrule
    \end{tabular}
    }
\end{table}